\documentclass{article}

\usepackage[preprint]{neurips_2025}

\usepackage[utf8]{inputenc} % allow utf-8 input
\usepackage[T1]{fontenc}    % use 8-bit T1 fonts
\usepackage{url}            % simple URL typesetting
\usepackage{booktabs}       % professional-quality tables
\usepackage{amsfonts}       % blackboard math symbols
\usepackage{nicefrac}       % compact symbols for 1/2, etc.
\usepackage{microtype}      % microtypography
\usepackage{xcolor}         % colors

\usepackage{graphicx} 
\usepackage{amsmath}
\usepackage{amssymb}
\usepackage{mathtools}
\usepackage{amsthm}
\usepackage{algpseudocode}
\usepackage{algorithm}
\usepackage{wrapfig}
\usepackage{array}

\usepackage[colorlinks,
            linkcolor=blue,
            anchorcolor=blue,
            citecolor=gray
            ]{hyperref}

\title{Efficient Online RL Fine Tuning with Offline Pre-trained Policy Only}

\author{%
  Wei Xiao \:
  Jiacheng Liu \:   
  Zifeng Zhuang \:  
  Runze Suo \\
  \textbf{Shangke Lyu\textsuperscript{*}  \: 
  Donglin Wang\thanks{Corresponding author.}}\\
  \normalfont 
  Westlake University\\
  }

\begin{document}

\maketitle

\begin{abstract}
% Firstly, the pessimism in Q-functions trained via offline reinforcement learning (RL), though crucial for ensuring policy optimization within the data support region during pre-training, could significantly hinder the efficiency of online fine-tuning.
% This occurs because these Q-functions systematically underestimate out-of-distribution state-action pairs, even when such exploration could yield higher rewards.
% Secondly, the absence of pre-trained Q-functions in scenarios where only a pre-trained policy is available, such as imitation learning (IL) pre-training, renders traditional offline-to-online RL methods less applicable.

% Conservatism is commonly used in offline reinforcement learning (RL) to ensure policy optimization within the data support region. 
% However, this conservatism leads to pessimistic Q-functions, which may hinder exploration when transitioning from the offline to the online setting. 
% Specifically, these Q-functions systematically underestimate state-action pairs beyond the offline dataset, limiting further exploration that could yield higher rewards. 

Improving the performance of pre-trained policies through online reinforcement learning (RL) is a critical yet challenging topic.
Existing online RL fine-tuning methods require continued training with offline pretrained Q-functions for stability and performance.
However, these offline pretrained Q-functions commonly underestimate state-action pairs beyond the offline dataset due to the conservatism in most offline RL methods, which hinders further exploration when transitioning from the offline to the online setting. 
Additionally, this requirement limits their applicability in scenarios where only pre-trained policies are available but pre-trained Q-functions are absent, such as in imitation learning (IL) pre-training.
To address these challenges, we propose a method for efficient online RL fine-tuning using solely the offline pre-trained policy, eliminating reliance on pre-trained Q-functions.
We introduce \textbf{PORL} (\textbf{P}olicy-\textbf{O}nly \textbf{R}einforcement \textbf{L}earning Fine-Tuning), which rapidly initializes the Q-function from scratch during the online phase to avoid detrimental pessimism.
Our method not only achieves competitive performance with advanced offline-to-online RL algorithms and online RL approaches that leverage data or policies prior, but also pioneers a new path for directly fine-tuning behavior cloning (BC) policies.
\end{abstract}

\section{Introduction}

% background (pertraining -> fine-tuning)
The paradigm of pre-training decision-making policies through offline RL \citep{levine2020offline, kostrikov2021offline, kostrikov2021offline1} and imitation learning \citep{wu2019behavior, zhao2023act, chi2023diffusion} has emerged as a cornerstone for scalable policy learning. These methods leverage pre-collected datasets or expert demonstrations to develop policies that perform well in diverse tasks.
Despite these advancements, pre-trained policies may fail to achieve optimal performance or adapt to new task variations due to the limited coverage, diversity, and quality of training data. Online RL fine-tuning offers a promising solution by improving policies through environmental interaction.

\begin{figure*}[ht]
    \centering
    \includegraphics[width=1\linewidth]{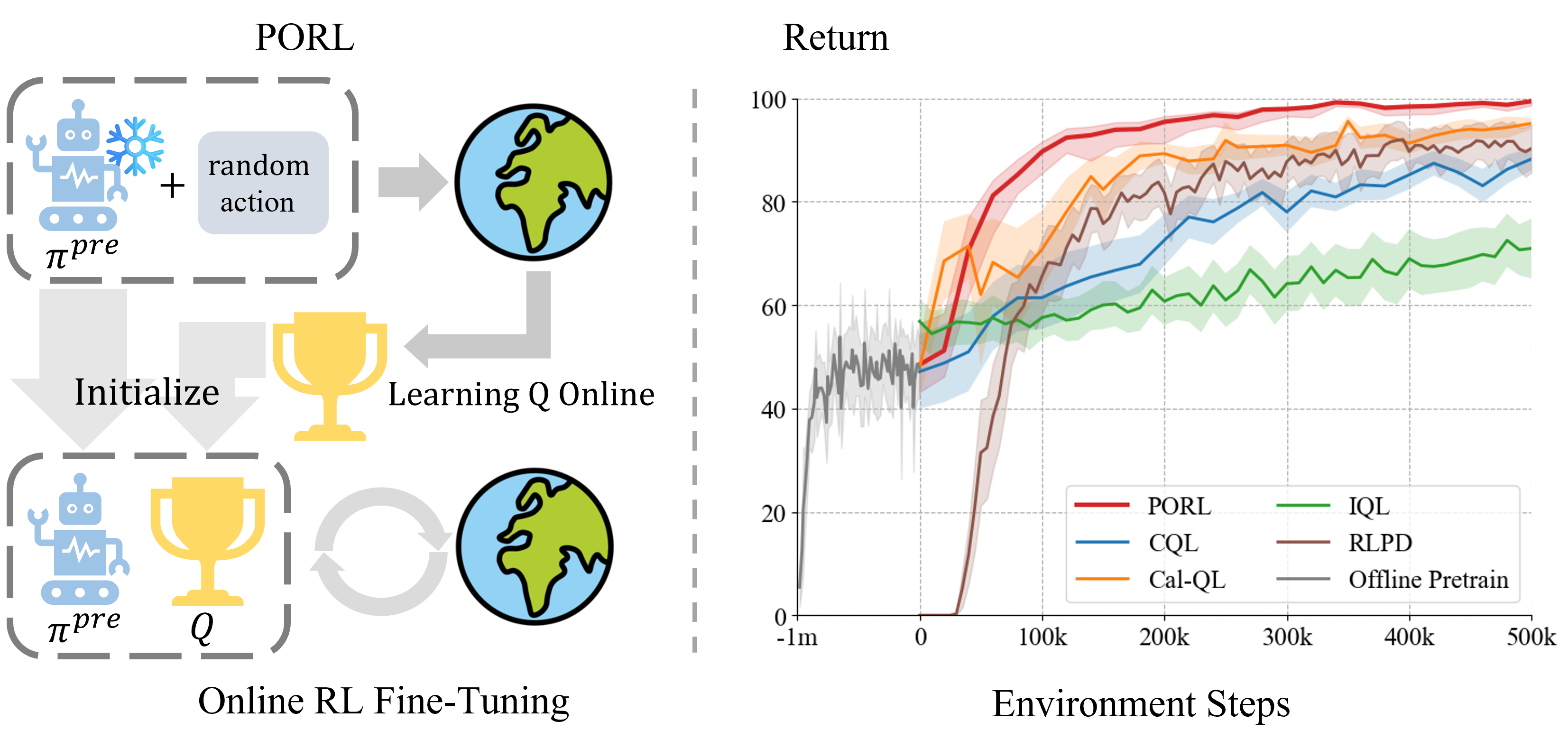}
    \vspace{-0.2in}
    \caption{Online RL fine-tuning with pre-trained policy only. The framework overview of PORL (left); Performance comparison on all antmaze tasks (right).}
    \label{fig:intro}
    \vspace{-0.1in}
\end{figure*}

% challenges (Pessimistic Q & absence of Q)
However, the transition from offline pre-training to online fine-tuning is non-trivial. 
% Pessimistic Q
Firstly, offline RL techniques including policy constraints \citep{fujimoto2019off, fujimoto2021minimalist} and value regularization \citep{kumar2020conservative} tend to induce biases in the learned Q-function, particularly toward underestimating the value of state-action pairs beyond the offline dataset. 
While this pessimism stabilizes offline training by mitigating overestimation errors, it becomes detrimental when transitioning to online fine-tuning. 
Specifically, the pre-trained Q-function with constraints or pessimism may assign systematically lower value estimation to novel transitions encountered during online interaction. 
Consequently, it could suppress exploration during the online stage, limiting policy extending to the unseen region with higher rewards.
% absence of Q
Secondly, previous offline-to-online reinforcement learning methods typically rely on both pre-trained policies and offline pre-trained Q-functions to facilitate fine-tuning. 
Such a dependency renders them inapplicable to imitation learning settings, where only a pre-trained policy is available, and any pre-trained Q-function is absent.

% problem setting (policy only)
To avoid the reliance on pre-trained Q-function, an immediate idea is: \textbf{Can we conduct efficient reinforcement learning fine-tuning only with the pre-trained policy?} Addressing this challenge would not only eliminate the pessimistic issue in offline Q-function but also bridge the critical gap in RL fine-tuning for scenarios where offline pre-trained Q-functions are unavailable.

% analysis
In this paper, we aim to build an online RL fine-tuning method solely with the offline pre-trained policy. We begin with an empirical analysis of RL fine-tuning under various Q-function initializations. Our study reveals that online tuning initialized with online-trained Q-functions achieves faster improvement and higher convergence scores compared to tuning initialized with offline pre-trained ones. Surprisingly, this result holds even when the evaluation performance of the online-trained Q-functions is initially poor.
From a Q-value estimation perspective, we observe that offline pre-trained Q-functions exhibit persistent underestimation for regions outside the offline data during online tuning. In contrast, online-trained Q-functions do not exhibit such biases, leading to more uniform value estimations.

The above finding inspires us to develop a method for learning a well-initialized Q-function online that avoids such biases of pre-trained Q-function.
To obtain an effective Q-function initialization for efficient online RL fine-tuning, we propose a simple yet practical approach: first freezing the pre-trained policy and using a sampling approach with randomness (e.g. epsilon greedy) to collect a few online interaction data to "warm up" the Q-function. 
We term this approach \textbf{PORL} (\textbf{P}olicy-\textbf{O}nly \textbf{R}einforcement \textbf{L}earning Fine-Tuning). 
Empirically, it achieves impressive results compared to offline-to-online RL methods and significantly outperforms online fine-tuning with random Q-function initialization.
Additionally, it enables direct fine-tuning of pretrained policies without requiring any pre-trained Q-function or offline dataset, a capability unattainable by prior methods.

In summary, our contributions are as follows:

\begin{itemize}
\item Analyze the impact of various Q-function initialization (random, offline-trained, and online-trained) in online RL fine-tuning and reveal the limitations of initialization with the offline-trained Q-function.
\item Introduce PORL, an online RL fine-tuning method which enables direct fine-tuning of imitation learning-pretrained policies without requiring any pre-trained Q-function or offline dataset.
\item Conduct extensive experiments and ablations to study the effectiveness of our approach. Empirically, it achieves competitive results compared to advanced offline-to-online RL methods and significantly outperforms online fine-tuning with random Q-function initialization.
\end{itemize}

\section{Problem Formulation}
We consider an infinite-horizon Markov Decision Process (MDP), $\mathcal{M} = \{\mathcal{S}, \mathcal{A}, \mathbf{P}, r, \gamma, \rho\}$, which is defined by the state space $\mathcal{S}$, the action space $\mathcal{A}$, the transition dynamics $\mathbf{P}(s' | s, a): \mathcal{S} \times \mathcal{A} \to \mathcal{P}(\mathcal{S})$, the reward function $r: \mathcal{S} \times \mathcal{A} \to \mathbb{R}$, the discount factor $\gamma \in [0, 1)$, and the initial state distribution $\rho : \mathcal{P}(\mathcal{S})$.

In this work, we focus on a setting where the sole available resource from offline training is a pre-trained policy  $\pi^\mathrm{pre}(a|s): \mathcal{S} \to \mathcal{P}(\mathcal{A})$, which has been derived using offline reinforcement learning or imitation learning methods. 
Unlike conventional offline-to-online RL approaches, where offline datasets $\mathcal{D}_\mathrm{off}$ and pre-trained value functions $Q^\mathrm{pre}$ are accessible to support online fine-tuning, our setting assumes no access to offline data or pre-trained critics. 
This creates a unique challenge, as existing offline-to-online RL algorithms heavily depend on these components to facilitate policy improvement.
The objective of our problem is to design an online fine-tuning algorithm that exclusively relies on the pre-trained policy $\pi^\mathrm{pre}(a|s)$ and interaction with the environment to maximize the expected discounted return:
\begin{align*}
\eta(\pi) = \mathbb{E}_{s_{t+1} \sim \mathbf{P}(\cdot | s_t, a_t), a_t \sim \pi(\cdot | s_t), s_0 \sim \rho}\left[\sum_{t=0}^{\infty} \gamma^t r(s_t, a_t)\right],
\end{align*}
where $\pi$ denotes the fine-tuned policy.

This problem setting emphasizes fine-tuning in the absence of offline data and pre-train initialized value functions, requiring methodologies that effectively leverage $\pi^\mathrm{pre}$.

\section{Q-function Initialization in Online RL Fine-Tuning}
\label{sec3}
\subsection{Randomized Q-function Initialization}
\begin{wrapfigure}{r}{0.6\linewidth}
\vspace{-0.25in}
  \centering
  \includegraphics[width=1\linewidth]{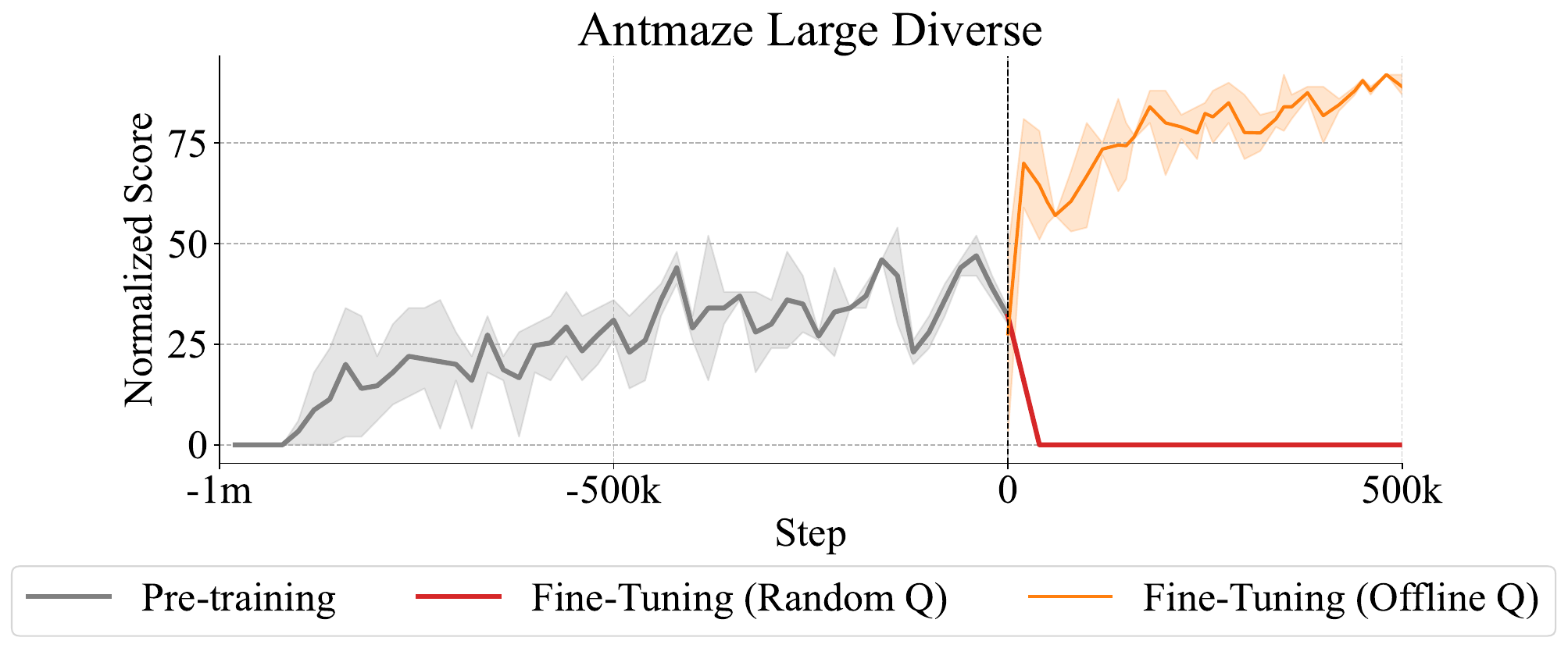} 
  \vspace{-0.25in}
  \caption{An example of actor-critic tuning with randomized and offline-trained (Cal-QL) Q-function initialization.}
  \label{fig:case}
  \vspace{-0.2in}
\end{wrapfigure}

Previous research \citep{uchendu2023jump, zhou2024efficient} together with our experiments in Figure~\ref{fig:case} indicate that when using only a pre-trained policy to initialize the actor in an actor-critic algorithm, the randomly initialized critic provides poor learning signals. 
This results in a degradation of the actor's performance, causing the initially well-performing policy to be forgotten.

\subsection{Offline Q-function vs. Online Q-Function Initialization}

To address the limitation of using randomly initialized critics in actor-critic algorithms and explore the impacts of Q-function initialization, we compare the effects of various Q-function initializations on online fine-tuning.
We first use Cal-QL \citep{nakamoto2024cal} for offline RL pretraining on the \texttt{halfcheetah-medium-v2} dataset, obtaining a pre-trained policy (\(\pi^{pre}\)) with a normalized score of 48 and a pre-trained Q-function (\(Q^{off}\)). 
For comparison, we extract three online-trained Q-functions (\(Q^{on}_{low}\), \(Q^{on}_{mid}\), and \(Q^{on}_{high}\)) from different stages of SAC \citep{haarnoja2018soft} training process. 
These three Q-functions correspond to online policies that achieved normalized scores of 12, 42, and 52 at extraction time.

\begin{figure}[htbp]
    \centering
    \includegraphics[width=1\linewidth]{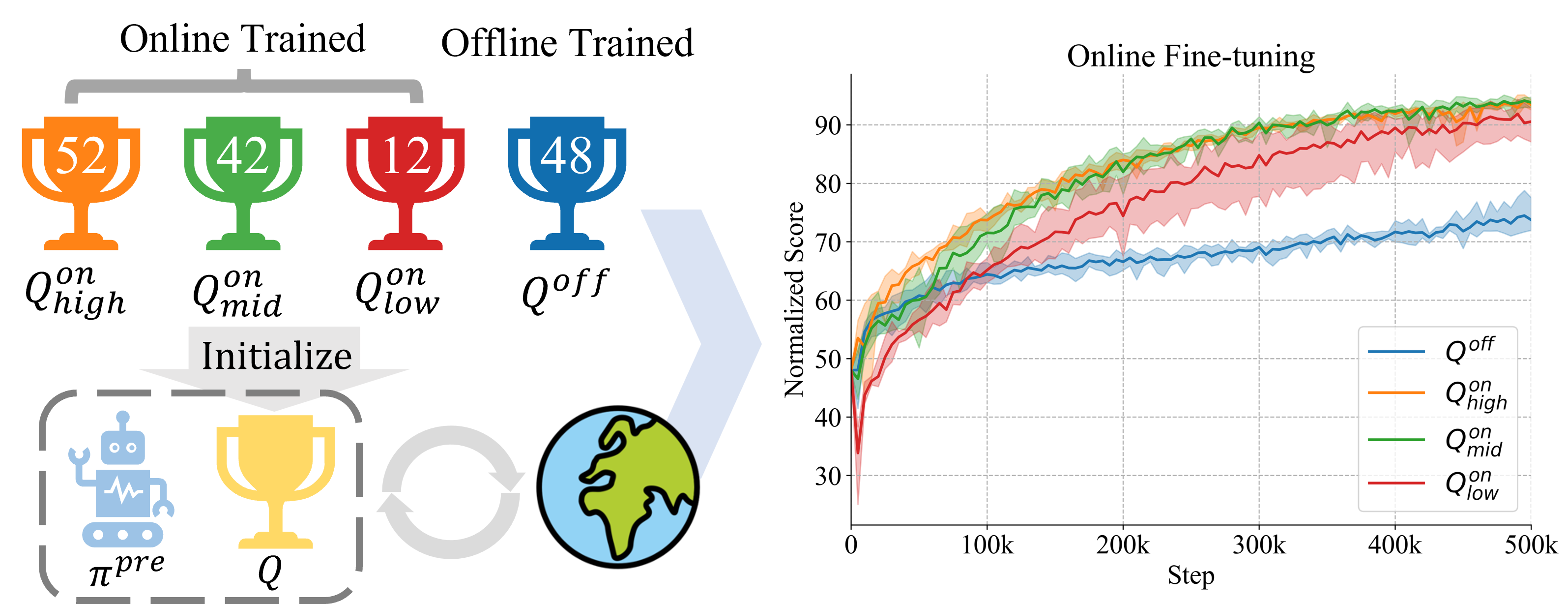}
    \vspace{-12pt}
    \caption{Comparison of online fine-tuning with various Q-function initializations.}
    \label{fig:different_Q}
\end{figure}

The results shown in Figure \ref{fig:different_Q} indicate that \textbf{all online fine-tuning initialized with $Q^{on}$ achieve faster improvement and higher convergence scores compared to that initialized with $Q^{off}$.}
Specifically,
$Q^{on}_{high}$ outperformed $Q^{off}$, demonstrating superior performance due to its higher initial score.
$Q^{on}_{mid}$, despite having a slightly lower initial score than $Q^{off}$, also achieved better results. 
Surprisingly, \(Q^{on}_{low}\), which starts with the lowest score, causes an initial performance drop due to the mismatch with \(\pi^{pre}\). 
However, it recovers rapidly during fine-tuning, eventually surpassing \(Q^{off}\).
These observations suggest a limitation in the effectiveness of $Q^{off}$ initialization for online training.

\subsection{Why Online Q-function Initialization is Better?}
The divergence between offline-pretrained and online-trained Q-functions arises from two key factors.

\begin{itemize}
\item Offline Q-functions (\(Q^{off}\)) are trained on fixed, narrow datasets with limited exploration \citep{fu2020d4rl}, while online counterparts (\(Q^{on}\)) learn from dynamically expanding data distributions; 
\item Offline RL inherently enforces conservatism via policy constraints or value regularization \citep{kumar2020conservative}, leading to pessimistic Q-value for out-of-distribution (OOD) state-action pairs. In contrast, online Q-functions freely estimate values without such constraints \citep{haarnoja2018soft}.
\end{itemize}

\begin{wrapfigure}{r}{0.6\linewidth}
\vspace{-0.25in}
  \centering
  \includegraphics[width=1\linewidth]{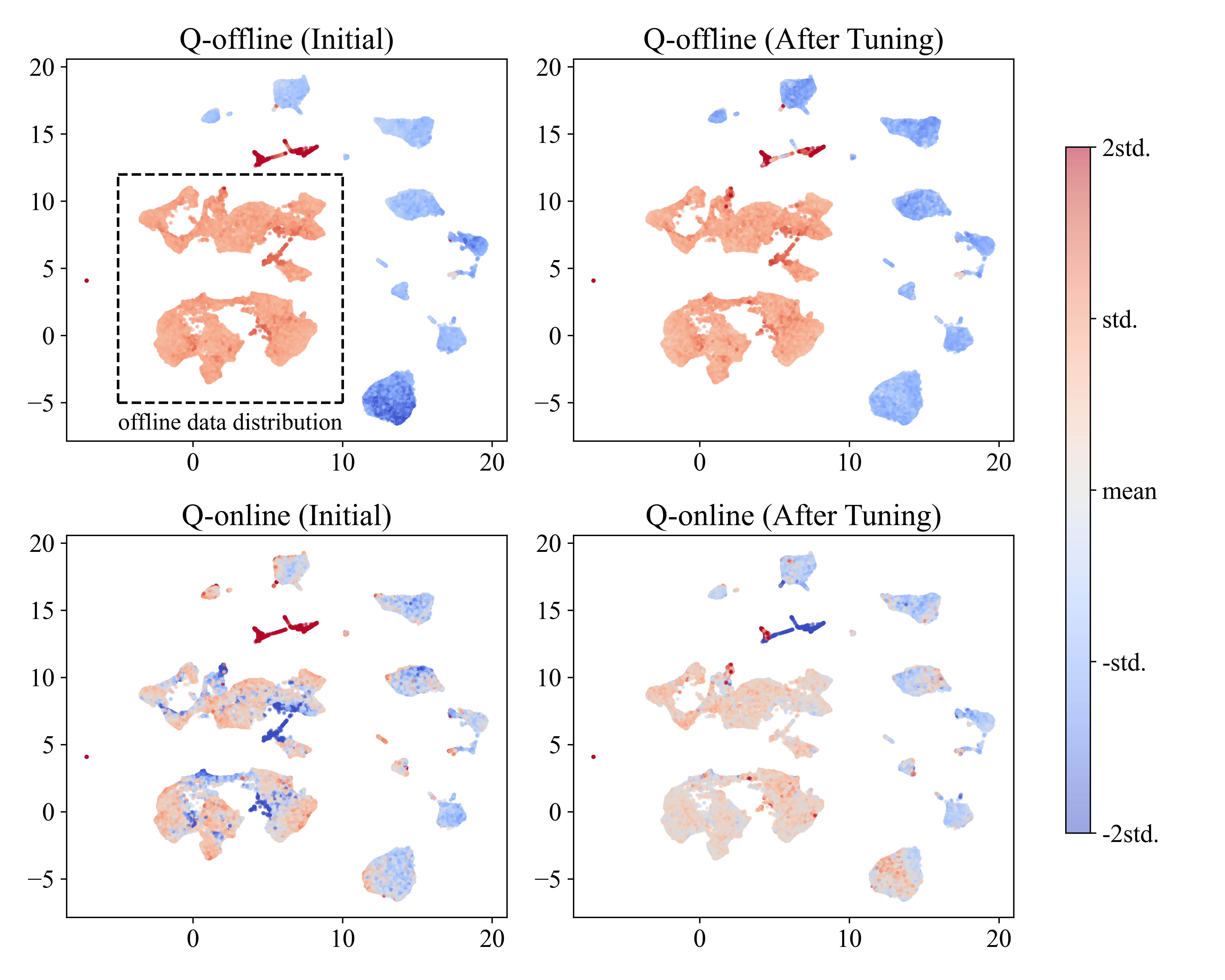}
  \caption{Q-Value estimation differences across various Q-function initialization before and after online fine-tuning. The Q-value bias labeled by Q-online or Q-offline and value reference before (left) and after (right) online training.}
  \label{fig:Q_diffs}
\vspace{-0.25in}
\end{wrapfigure}

The visualizations in Figure \ref{fig:Q_diffs} illustrate the Q-value estimation differences between Q-offline ($Q^{off}$) and Q-online ($Q^{on}_{low}$) initialization, helping us understand the results of these two distinct factors.
We first visualize the distribution of state-action pairs from the offline dataset and online fine-tuning collected to 2D plane via UMAP~\citep{mcinnes2018umap} (Uniform Manifold Approximation and Projection) and label these state-action pairs' Q-value bias. 
The bias is the Q-value minus the reference value, which is labeled by the Q-function from running SAC online to convergence.
In general, the \textcolor{blue!70!white}{cooler} the labeled color indicates that these state-action pairs are more underestimated.

Due to the conservative constraints during the offline training phase, Q-offline tends to underestimate the Q-values of state-action pairs which out of the offline data distribution (OOD). Even after 500k steps of online fine-tuning, it remains the pessimistic estimation for the OOD region, indicating a persistent bias.
In contrast, Q-online treats state-action pairs from both the offline data and the OOD region more equally due to the inherent uncertainty in online sampling. This results in a more uniform Q-value distribution without significant underestimation in Q-offline. 
Such uniformity in Q-value estimation is more conducive to further exploration in online fine-tuning.
The resulting state distribution in App.~\ref{a1} indicates that the policy tuning with Q-online expands to more high-value region.

In summary, the more uniform Q-value estimation is advantageous for online fine-tuning. 
The reason for this uniformity is that the training data for Q-online comes from the stochastic nature of online sampling. 
This insight suggests that employing sampling methods with randomness to collect online data for Q-function learning may be beneficial in online fine-tuning.

\section{PORL: Policy-Only Reinforcement Learning Fine Tuning}
Inspired by the aforementioned observations, we hypothesize that learning a Q-function initialization through online interactions before policy training could contribute to more efficient online RL fine-tuning. 
Accordingly, we introduce a pre-sampling stage, where a few online interactions are used to rapidly learn a Q-function initialization. 
This stage aims to obtain a wider data distribution while maintaining alignment with the pre-trained policy’s data distribution. To facilitate this, we employ the \textbf{Epsilon-Greedy} sampling strategy. Other randomness-based sampling methods could also serve as promising alternatives worth exploring.
In epsilon-greedy, the agent selects actions from the pre-trained policy (exploitation) with probability \(1 - \epsilon\), where \(\epsilon\) is a small positive value (e.g., 0.1), and with probability \(\epsilon\), it selects a random action (exploration). This approach introduces controlled randomness, promoting exploration to ensure the collection of diverse interactions and guaranteeing the distribution does not deviate from the behavior of policy, which helps learn a more uniform and effective Q-function initialization.

\begin{algorithm}[h]
\caption{PORL}\label{alg:pseudocode}
\begin{algorithmic}[h]

\Require Pre-trained policy $\pi$, Randomly initialized $Q$, Update-to-data ratio (UTD) $M$, Replay buffer $\mathcal{D} \leftarrow \emptyset$, Pre-sample steps $T$, Max Environment Steps $N$.

\State \color{red!70!black} Pre-sample Stage:

\For{step = 1 to $T$}
    \State $(s, a, s', r) \gets$ $\mathrm{EpsilonGreedy}(\pi, env)$
    \State $\mathcal{D} \gets \mathcal{D} \cup \{(s, a, s', r)\}$
    \State $\mathrm{Batch}_1, \mathrm{Batch}_2, ..., \mathrm{Batch}_{M} \sim \mathcal{D}$
    \State $Q \gets \mathrm{TemporalDifferenceUpdate}(Q, \mathrm{Batch}_i)$ for $M$ times
\EndFor

\State \color{black} SAC Update:

\For{step = $T+1$ to $N$}
    \State $(s, a, s', r) \gets$ $\mathrm{Interact}(\pi, env)$
    \State $\mathcal{D} \gets \mathcal{D} \cup \{(s, a, s', r)\}$
    
    \State $\mathrm{Batch}_1, \mathrm{Batch}_2, ..., \mathrm{Batch}_{M} \sim \mathcal{D}$
    \State $Q \gets \mathrm{TemporalDifferenceUpdate}(Q, \mathrm{Batch}_i)$ for $M$ times

    \State $\pi \gets \mathrm{PolicyGradient}(\pi, \mathrm{Batch}_1 \cup \cdot \cdot \cdot \cup \mathrm{Batch}_{M})$
\EndFor

\end{algorithmic}
\end{algorithm}

The algorithm pipeline is shown in pseudo-code \ref{alg:pseudocode}, highlighting in \textcolor{red!70!black}{Red} elements that are the main changes to our approach.
The omitted portions of the pseudo-code adhere to the general framework of SAC.

To accelerate Q-function learning, we adopt a \textbf{High Update-To-Data (UTD) ratio}, which increases the frequency of critic updates per online interaction. Specifically, we set the UTD ratio to 16 across all tasks, enabling the critic to learn efficiently from the limited interactions available during the pre-sample stage.
To mitigate potential Q-value overestimation caused by frequent updates, we incorporate two complementary techniques: \textbf{Layer Normalization} and \textbf{Q-Ensemble}. Layer Normalization is applied to the critic networks to prevent unreasonable generalization and stabilize the learning process.
Additionally, we enhance the robustness of Q-value estimation by employing Q-Ensemble, where multiple Q networks are trained in parallel. In our work, we use an ensemble of 10 Q networks, which ensures a more stable and reliable Q-function estimate, especially during the early stages of online fine-tuning.
These three techniques are commonly used together in current work on offline RL \citep{an2021uncertainty, yue2024understanding}, online RL \citep{chen2021randomized, ball2023efficient}, and offline-to-online RL \citep{zhou2024efficient, zhang2024perspective, zhao2024enoto, feng2024suf}, and be effective in improving the estimation and stability of the Q-value.

\section{Experiments}
\subsection{Main Results}
Our experimental evaluation aims to study how well PORL can conduct online RL fine-tuning with pre-trained policies only.
We experiment on Antmaze, Adroit, and Kitchen tasks from D4RL \cite{fu2020d4rl} and the RLbench \cite{james2019rlbench} tasks (see App.~\ref{benchmark} for benchmark details).
The compared methods include the advanced offline, offline-to-online RL, and online RL algorithms (see App.~\ref{baseline} for baseline details and implementation).

% policy-only setting
\begin{figure*}[htbp]
    \centering
    \includegraphics[width=1.0\linewidth]{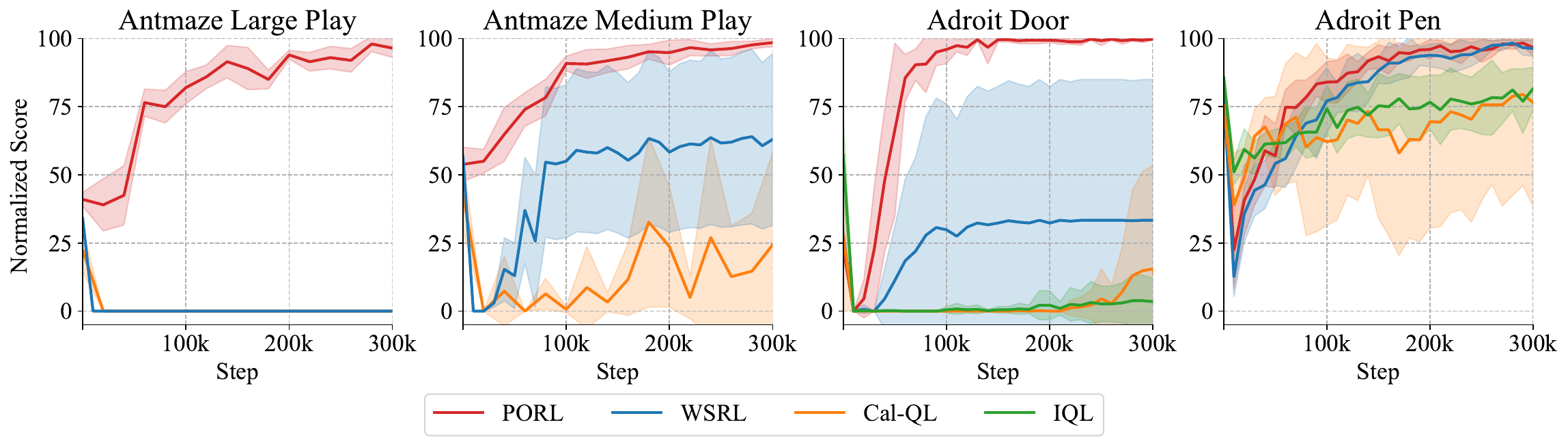}
    \vspace{-0.2in}
    \caption{Comparative experiments on online fine-tuning with \textbf{pre-trained policies initialized only}.}
    \label{fig:policy_only}
    \vspace{-0.1in}
\end{figure*}

Fig.~\ref{fig:policy_only} compares PORL with baselines which only initialized with the pre-trained policy.
The commonly used offline-to-online method performs very poorly when only the actor is initialized.
For these methods, it is crucial to utilize a Q-function trained by offline RL as a starting point, while PORL overcomes this limitation.

% normal-setting
\begin{figure*}[htbp]
    \centering
    \includegraphics[width=1\linewidth]{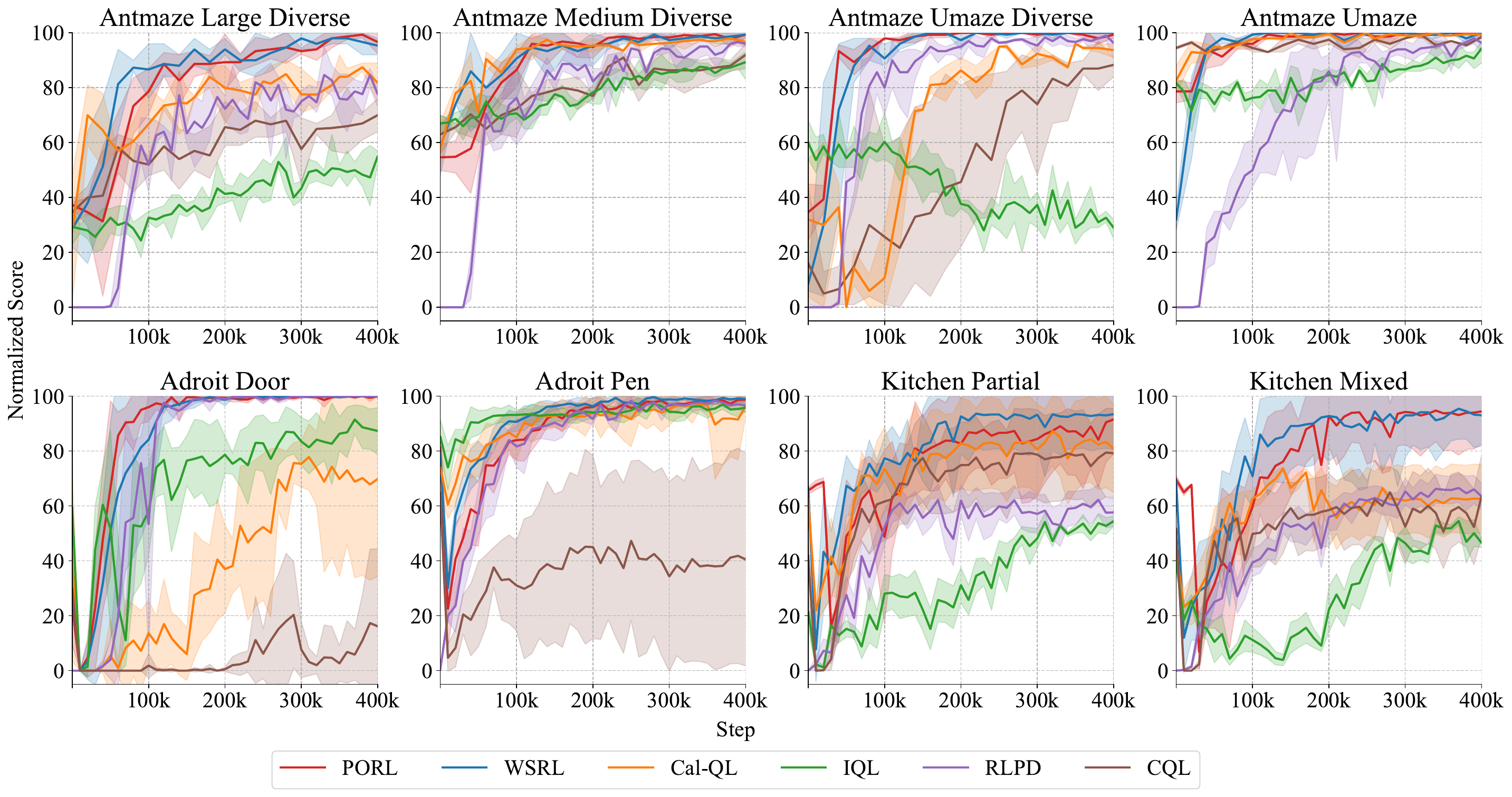}
    \vspace{-0.2in}
    \caption{Comparative experiments on offline-to-online tasks.}
    \label{fig:off2on}
    \vspace{-0.1in}
\end{figure*}

Fig.~\ref{fig:off2on} reveals that PORL achieves superior asymptotic performance compared to baseline methods.
CQL, IQL, and Cal-QL exhibit suboptimal convergence due to their conservative Q-function initialization inherited from offline RL, which systematically underestimates novel state-action pairs during the online stage.
RLPD, despite leveraging offline data through symmetric sampling, suffers from slow convergence due to policy initialization from scratch and achieves suboptimal results.
PORL eliminates reliance on pre-trained Q-function and achieves competitive performance with advanced methods, while WSRL still requires pre-trained Q-function initialization, limiting its applicability to imitation learning scenarios.
We also evaluate the performance of WSRL without pre-trained Q-function initialization (see App.~\ref{WSRL}), which performs poorly than PORL in most tasks.

\begin{figure*}[htbp]
    \centering
    \includegraphics[width=1\linewidth]{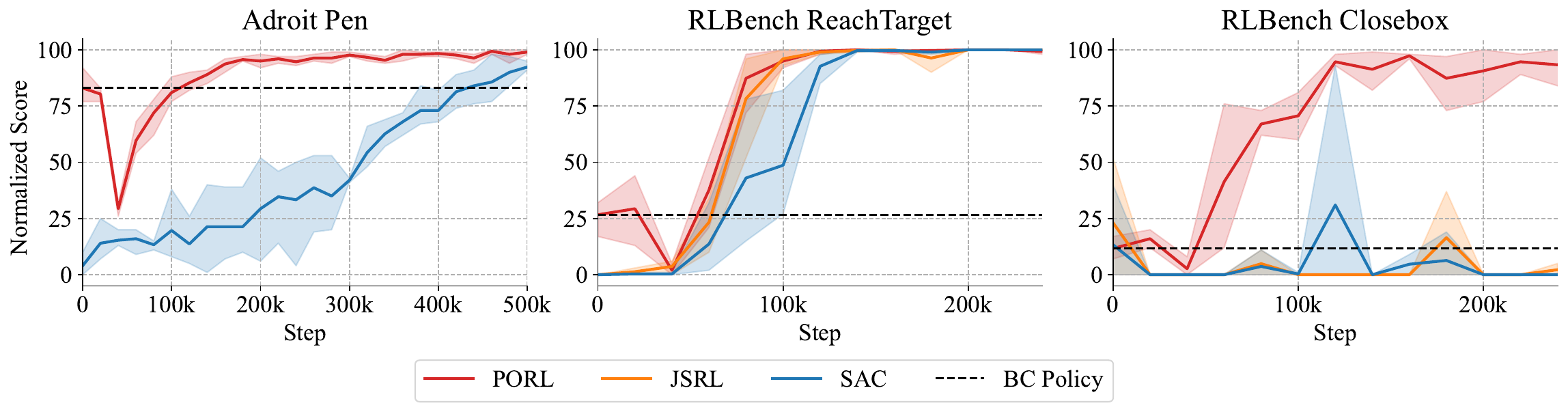}
    \vspace{-0.2in}
    \caption{Comparative experiments on online fine-tuning with pre-trained policies trained by Behavioral Cloning.}
    \label{fig:bc_setting}
    \vspace{-0.1in}
\end{figure*}

Fig.~\ref{fig:bc_setting} evaluates performance on three sparse-reward manipulation tasks, where the reward is binary (+1 for success, 0 for failure).
We initialize PORL with BC-trained policies  (dashed lines in the figure indicate the performance of these BC policies) and set these policies as the guide policies of JSRL.
PORL recovers rapidly from initial performance drops caused by critic-actor misalignment and converges to optimal success rates, outperforming baselines.
Notably, in the most challenging minipulation task \texttt{RLbench-Closebox}, JSRL and vanilla SAC fail, while PORL performs well.

In summary, our experiments demonstrate that PORL effectively improves the performance of both offline RL-trained and BC-trained policies.
Compared to existing methods, it achieves this with minimal dependencies, requiring only a pre-trained policy that is inherently available in any fine-tuning scenario.

\subsection{Ablation Studies}
We conducted a series of experiments to evaluate the effects of key design choices in our proposed method.

\begin{figure}[h]
    \centering
    \includegraphics[width=1\linewidth]{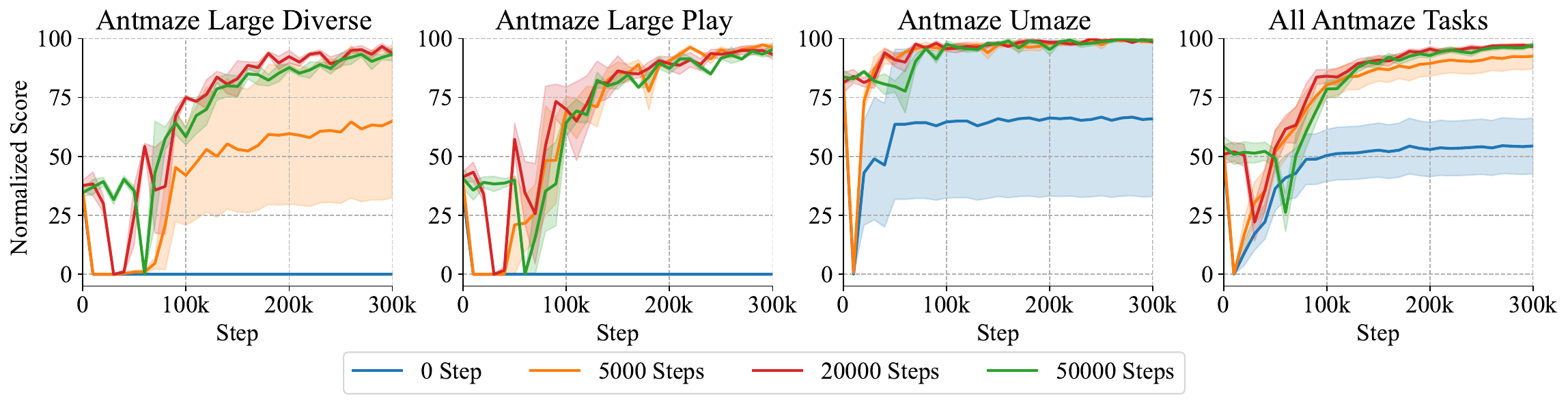}
    \vspace{-0.2in}
    \caption{Impact of pre-sample steps, evaluated every 10k steps and UTD=4.}
    \label{fig:steps}
        \vspace{-0.1in}
\end{figure}

\textbf{Impact of Pre-sample Steps}
We find that the number of pre-sample steps is critical for the Q-function initialization. 
Fig.~\ref{fig:steps} illustrates the impact of varying pre-sample steps
When the pre-sample steps are set to zero, the agent is equivalent to that initialized with a random critic and a pre-trained actor, which leads to poor performance and completely fails in challenging tasks.
At 5k pre-sample steps, the initialization becomes more effective, but it is still insufficient for optimal fine-tuning. 
Setting of pre-sample steps depends on the complexity of task and the capability of pre-trained policy.
When the task is more complex, more steps are needed because the pre-sampled data need to cover a certain area of the distribution.

\begin{figure}[h]
    \centering
    \includegraphics[width=1\linewidth]{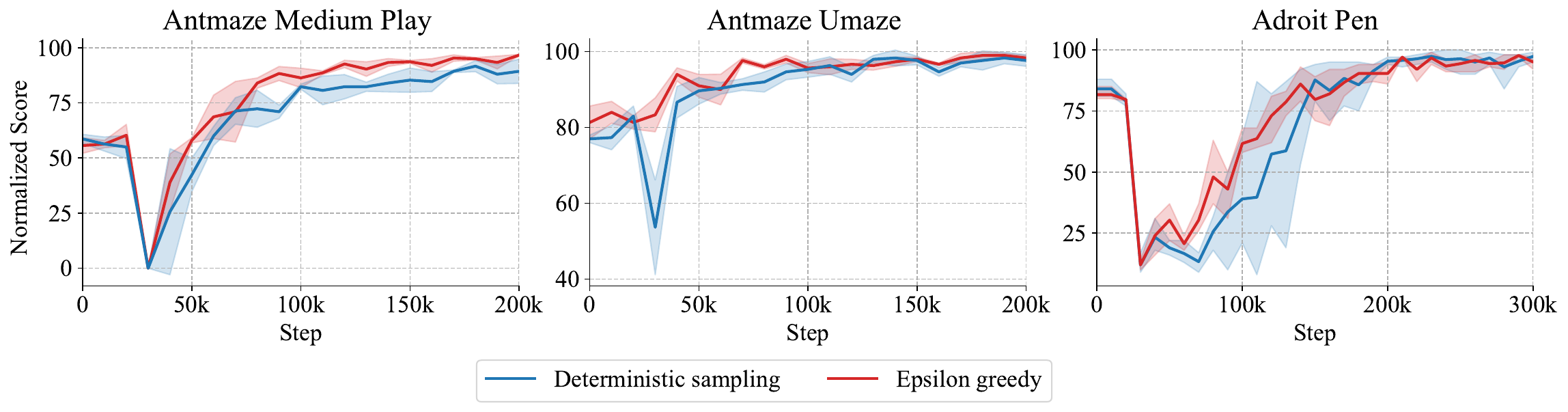}
    \vspace{-0.2in}
    \caption{Sampling methods for the pre-sample stage, evaluated every 10k steps and UTD=4.}
    \label{fig:epsilon}
        \vspace{-0.1in}
\end{figure}

\textbf{Sampling Methods for the Pre-sample Stage}
We explored different sampling strategies for the pre-sample stage, comparing epsilon-greedy with deterministic sampling. 
The results in Fig.~\ref{fig:epsilon} indicate that epsilon-greedy sampling is advantageous when the behavior distribution of the pre-trained policy is narrow.
This narrowness is generally a result of limited coverage in the pre-training data. However, when the data coverage is sufficiently diverse, the performance of these two sampling strategies is close.
By introducing controlled randomness, the sampling strategy expands the diversity of collected transitions, facilitating the learning of a more uniform value estimate of Q-function initialization.

\subsection{How to Mitigate Initial Performance Drops}
In tasks where the pre-trained policies are weak, we observed a performance drop at the start of training but rapidly recover, particularly in harder tasks.
The initial performance drop stems from misalignment between the actor and critic during their mutual adaptation phase: 
The actor's rapid policy changes destabilize the critic's value estimates, while the evolving feedback of the critic confounds the actor's optimization.
We explored two strategies, introducing a Behavioral Cloning (BC) term to constrain the actor's policy updates and increasing UTD to accelerate critic updates for this issue.

\textbf{Behavioral Cloning in Policy Learning}
From the actor's perspective, we introduce a BC term by adding a regularization component that encourages the actor to maintain behavior similar to the replay buffer. 
The BC term is weighted by \(\beta\), to balance the influence of the behavior cloning loss and the SAC loss. 
The policy training loss is modified as follows:
\begin{align*}
\mathcal{L}_{\text{BC-SAC}} = \mathbb{E}_{s_t, a_t \sim \text{batch}} \left[ \beta \cdot \log \pi( a_t | s_t) \right. + (1 - \beta) \left. \left( Q(s_t, \pi( \cdot | s_t)) - \alpha \log \pi( \cdot | s_t) \right) \right].
\end{align*}

\begin{wrapfigure}{r}{0.55\linewidth}
    \centering
    \includegraphics[width=1\linewidth]{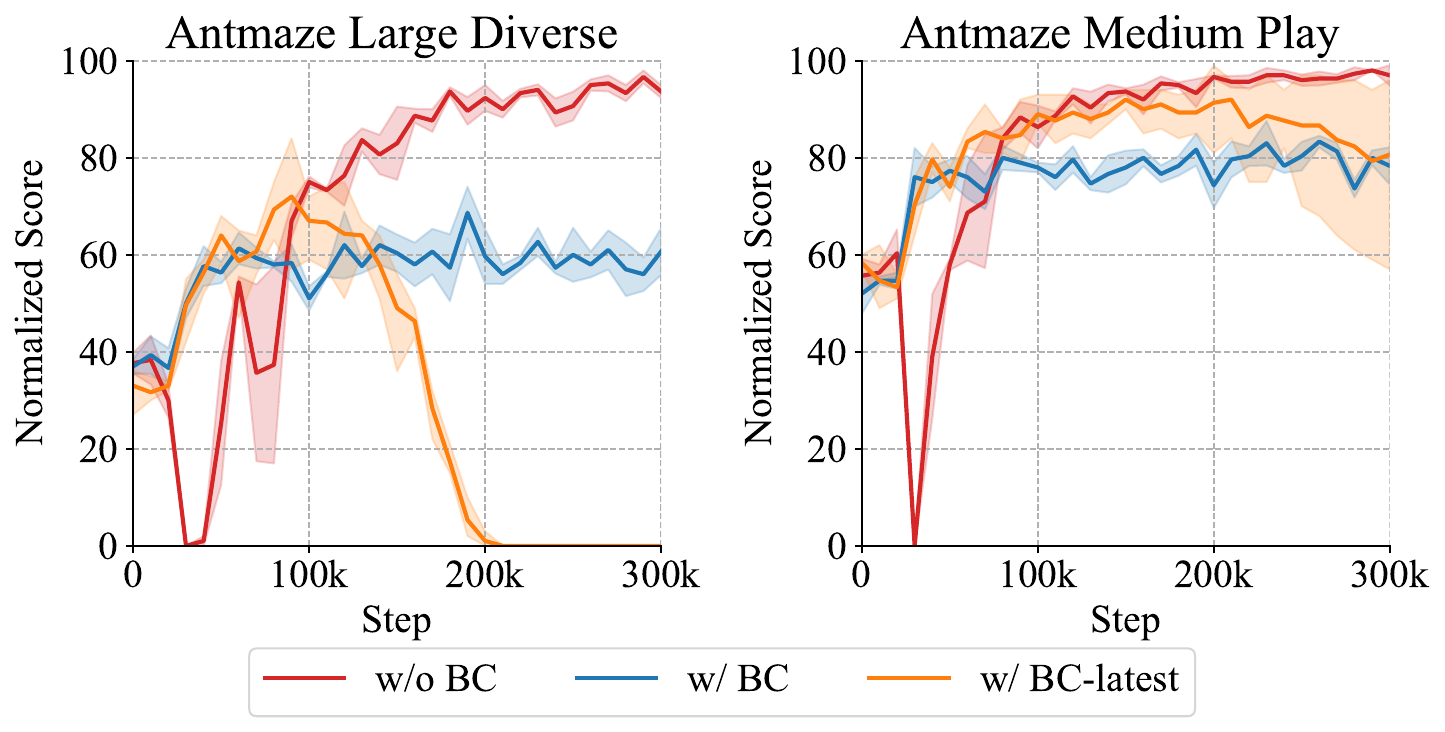}
    \caption{Adding a behavioral cloning (BC) term to the policy update objective, evaluated every 10k steps and UTD=4.}
    \label{fig:bc}
\end{wrapfigure}

The BC term encourages the actor to stay closer to the pre-trained behavior, thereby slowing the drastic changes in the policy during the early training stages. 
This regularization term effectively mitigates the initial performance degradation. 

As shown in Fig.~\ref{fig:bc}, we observed that while this adjustment eliminates the initial drop in performance, it could limit asymptotic performance by restricting the exploration of the unseen region.
In addition, we tried to clone the latest 20k transitions in the replay buffer (w/ BC-latest).
This approach achieves a faster rise in the early stages compared to cloning all transitions in the replay buffer (w/ BC), but crashes in the later stages of training.

\textbf{Higher UTD}
From the critic's perspective, we found that the higher UTD ratio effectively mitigates the initial performance drop.
By accelerating the critic’s learning process, we ensure that the critic provides more stable feedback to the actor at each step, thus reducing the fluctuations in the actor’s behavior.

\begin{figure}[htbp]
    \centering
    \includegraphics[width=1\linewidth]{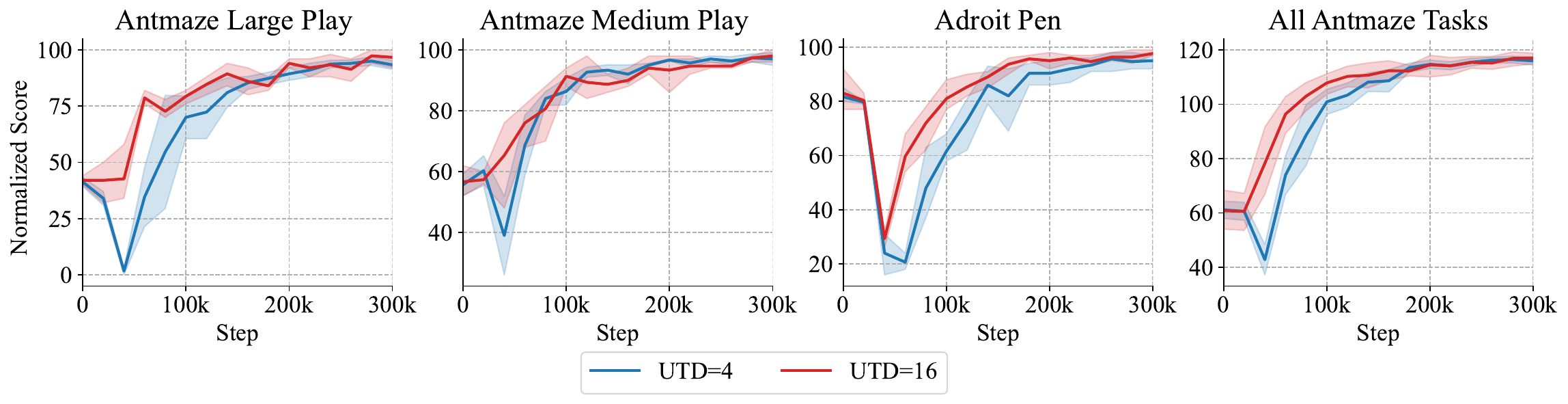}
    \vspace{-0.2in}
    \caption{Updating critic's parameters with higher UTD, evaluated every 20k steps.}
    \label{fig:utd}
        \vspace{-0.1in}
\end{figure}

As shown in Fig.~\ref{fig:utd}, we compared the effects of setting the UTD ratio to 4 versus 16. 
Although this adjustment helps stabilize the training in the early phase, it does not fully eliminate the performance drop, as the actor still needs time to adapt to the evolving critic.
The higher UTD ratio accelerates convergence while maintaining similar asymptotic results.  
However, this comes at increased computational costs per environment step, creating a trade-off between training cost and sample efficiency.  

\section{Related Works}
\subsection{Policy Pre-training}
Pre-training policies using a large-scale data-driven paradigm is a central challenge in decision-making research. Offline RL and imitation learning are two mainstream approaches for this task.

\textbf{Offline RL} extracts policies from large amounts of variable quality data with reward labels \citep{levine2020offline}.
\citet{wu2019behavior, fujimoto2021minimalist, kostrikov2021offline1, kumar2020conservative} focus on mitigating distributional shift issues between offline data and online environments by employing constraints or regularization to ensure stable policy improvement.
\citet{chen2021decision, ajay2022conditional, zhuang2024reinformer, lee2022multi, zheng2022online, wu2023elastic} treat RL as a sequence modeling problem and train an autoregressive model using offline dataset.

\textbf{Imitation learning}, particularly Behavioral Cloning (BC), trains policies to mimic expert actions through supervised learning \citep{pomerleau1988alvinn, zheng2022imitation, liu2023ceil}. 
These approaches rely on high-quality expert demonstrations without reward signals.
In robotic learning \citep{robot_il}, \citet{chi2023diffusion, shafiullah2022behavior, kim2024openvla, black2024pi0} have achieved substantial progress in pre-training policies for tasks such as manipulation, navigation, and control by utilizing large datasets or expert trajectories.

\subsection{Online RL Fine-Tuning}
Transitioning pre-trained policies to online environments introduces unique challenges.
Offline-to-online RL methods leverage pre-trained policy, Q-function, and offline dataset to run online RL efficiently. 
Initial attempt \citep{kostrikov2021offline, kostrikov2021offline1, nair2020awac, wu2022supported} focused on direct fine-tuning following offline training often suffered from excessive conservatism, leading to a lack of sample efficiency.
\citet{nakamoto2024cal, Chen2023DCACRU, feng2024suf, zhang2024improvingofflinetoonlinereinforcementlearning} mitigates this issue by calibrating the scale of Q-values or dynamically adjusting policy constraints. 
To explore the causes of pessimistic Q-functions, \citet{zhang2024perspective} proposes a novel perspective focusing on Q-value estimation bias.
\citet{ball2023efficient, song2022hybrid} integrate offline data into the online learning process, leveraging offline datasets to improve adaptation efficiency and reduce exploration needs.
\citet{uchendu2023jump, yuan2024policy} also begin with a pre-trained policy but treats it as a fixed guide, learning a separate exploration policy. While effective, they does not address scenarios where directly tuning the pre-trained policy is necessary.
Existing methods either depend on pre-trained critics \citep{nakamoto2024cal, Chen2023DCACRU, zhang2024perspective} or preclude direct policy adaptation \citep{uchendu2023jump, yuan2024policy}. 
Compared with them, our method bridges this gap by enabling direct fine-tuning of pre-trained policies—whether from offline RL or imitation learning—without requiring pre-trained Q-functions or offline data.

\section{Conclusion}
In this work, we investigate the feasibility of online RL fine-tuning using solely pre-trained policies.
This setting is crucial as it not only circumvents potential biases inherent in pre-trained Q-functions but also applies to scenarios where only pre-trained policies are available.
Consequently, previous offline-to-online RL algorithms, which rely on pre-trained Q-functions, fall short in this context.
We propose PORL, a minimalist yet effective approach for policy-only fine-tuning.
It demonstrates the challenges of this paradigm and establishes a foundation for future research.
In future work, we will investigate the fine-tuning of diverse model architectures of policies, within the policy-only setting.
With the rise of large-scale pre-trained policies, we believe that enhancing their performance and generalization through the RL paradigm represents a promising direction.

%\section*{References}

\bibliography{references}
\bibliographystyle{plainnat}

%%%%%%%%%%%%%%%%%%%%%%%%%%%%%%%%%%%%%%%%%%%%%%%%%%%%%%%%%%%%

% \appendix
\newpage 

\section*{Appendix}
\section{Future Directions and Limitations}
\label{limitation}
\begin{itemize}
    \item Q-function Evaluation: exploration on various Q-functions provides initial motivation and insights for our method. Future works could investigate how to evaluate the dynamics between actor-critic tuning. 
    \item Large model or Diffusion/Transformer model: As with previous approaches, our method still studies MLP-based policy networks. With the development of large-scale imitative learning, diffusion/transformer-based policy networks with large parameters have emerged challenges in online RL fine-tuning. How to use online RL to fine-tune these policy networks is a promising research direction.
    \item Real-world online fine-tuning: the cost of online interaction in a simulator is relatively inexpensive, however, the opposite is true in the real world. In some domains like robotics, the ultimate goal is performance in real-world tasks. Future works could explore the further enhancement of simulator-pretrained policies in the real world through online fine-tuning.
\end{itemize}

\section{The resulting state distributions of various Q-function initialization}\label{a1}
We further investigate the differences in the distribution of sampled states in various combinations during online fine-tuning.
The state distribution is visualized by UMAP.
Each data point represents the state and the value of the corresponding state-action pair, which is labeled by the Q-function from running SAC online to convergence.
\begin{figure*}[h]
    \centering
    \includegraphics[width=1\linewidth]{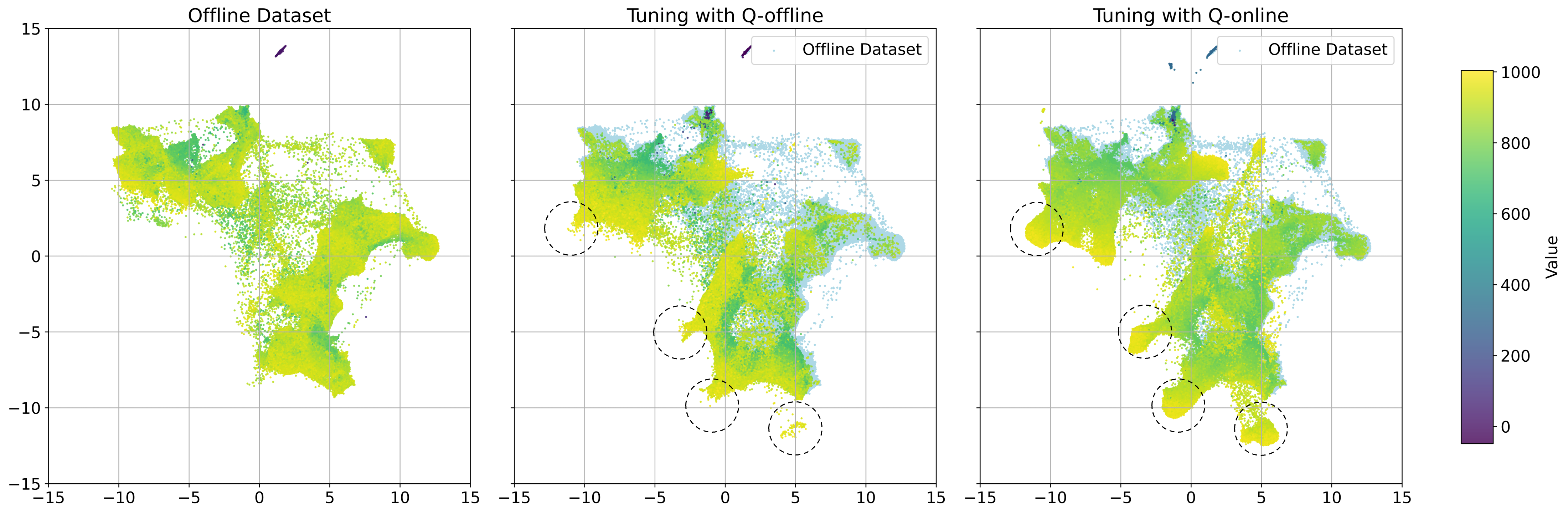}
    \caption{The resulting state distributions of various Q-function initialization and values of state-action pairs.}
    \label{fig:distribution}
\end{figure*}

A distribution shift exists between the offline dataset and the online data.
The distribution of data collected by the $Q^{off}$-fine-tuned policy is closer to the offline data distribution.
In contrast, the data collected using the $Q^{on}$-fine-tuned policy is more distant from the offline data distribution, potentially leading to the collection of state-action pairs with higher values not present in the offline dataset.

\section{Comparison with Warm-Start RL}
\label{WSRL}
\begin{figure}[htbp]
    \centering
    \includegraphics[width=1\linewidth]{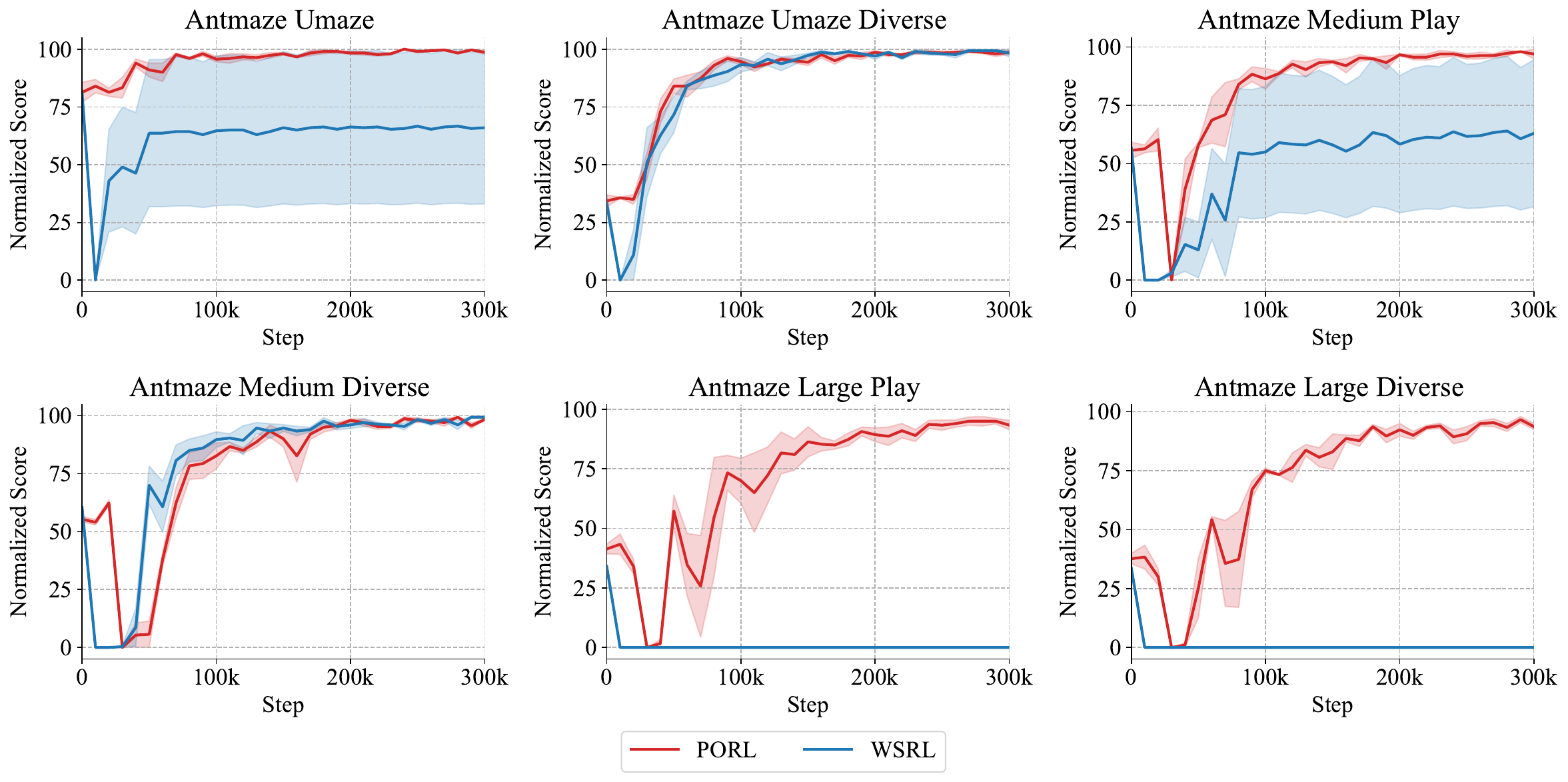}
    \caption{Comparison between PORL and WSRL with pre-trained policies initialization only (UTD=4).}
    \label{fig:wsrl}
\end{figure}

Warm-Start RL (WSRL) \cite{zhou2024efficient} is a recent offline-to-online RL algorithm that proposed no-retaining offline data during online tuning.
For scope and requirements, WSRL relies on a pre-trained Q-function for initialization, while PORL eliminates this dependency. This makes PORL more broadly applicable (e.g., to imitation-learned policies).
As shown in Fig.~\ref{fig:wsrl}, WSRL fails to achieve competitive performance without a pre-trained Q-function, whereas PORL excels in such scenarios.

WSRL's warmup stage aims to mitigate the offline-to-online distribution shift. In the Warmup Stage, the policy and Q-function are frozen, and the replay buffer is initialized by the pre-trained policy to interact with the environment.
PORL’s pre-sample stage is to acquire data in the online stage to quickly learn an appropriate Q-function and thus improve efficiency. Through the online interaction from the pre-trained policy, we can acquire a certain quality of online data to learn a suitable Q-function initialization, which makes our approach more efficient than directly following the offline pre-trained Q-function or a randomly initialized Q-function for policy updating in online scenarios.
While both methods omit offline data retention, PORL’s Q-learning independence avoids WSRL’s limitations when pre-trained Q-functions are suboptimal or unavailable

\section{Comparison with Jump-Start RL}
\label{JSRL}

\begin{figure}[htbp]
    \centering
    \includegraphics[width=1\linewidth]{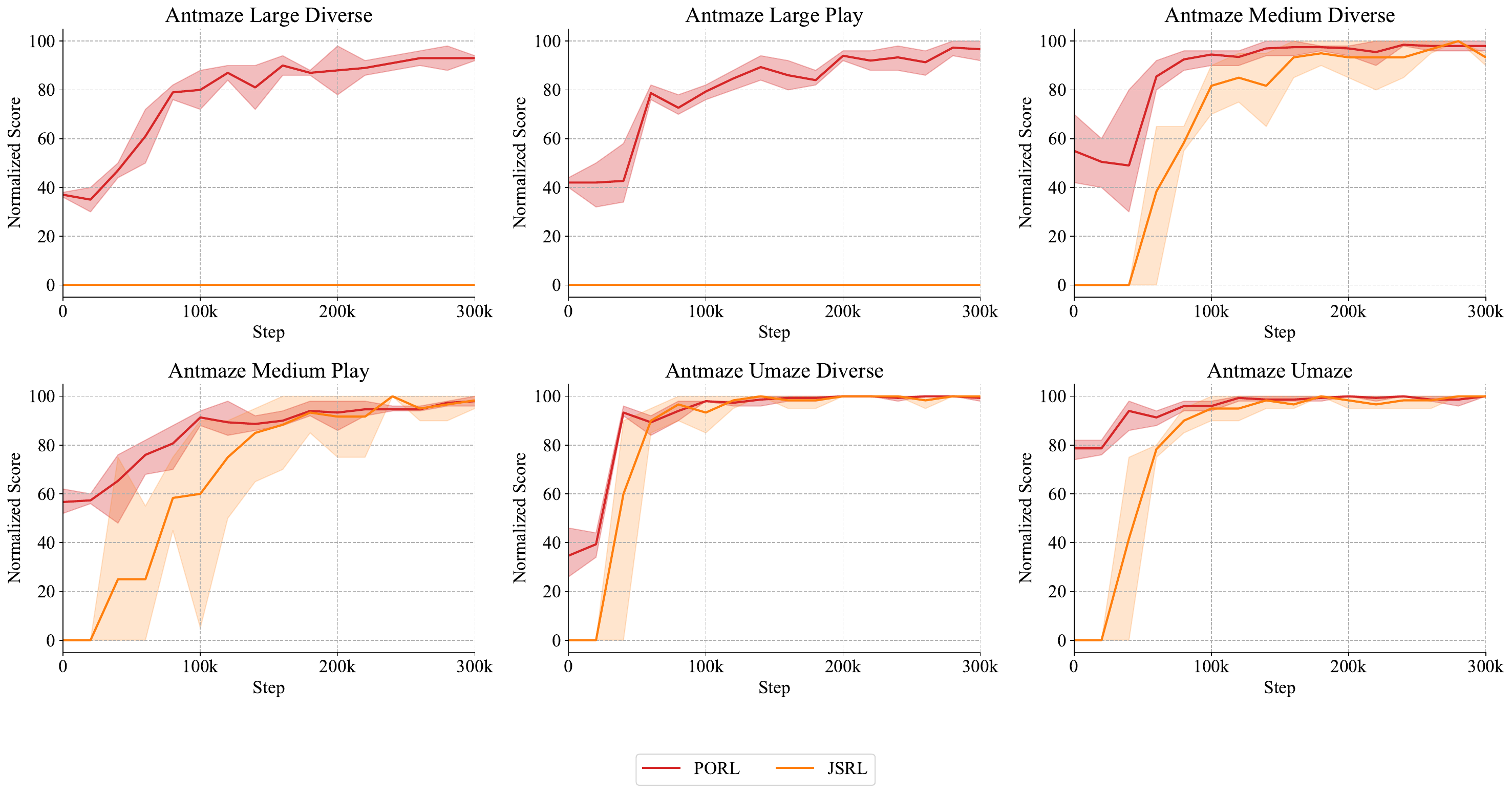}
    \caption{Comparison between PORL (UTD=16) and JSRL.}
    \label{fig:jsrl}
\end{figure}

Jump-Start RL (JSRL) \cite{uchendu2023jump} is a curriculum learning-based online RL algorithm that utilizes a pre-trained policy as a prior.
A closely related problem setting is explored in Jump-start reinforcement learning (JSRL) \cite{uchendu2023jump}, which also begins with a pre-trained policy. 
However, JSRL treats the pre-trained policy as a fixed guide to collect training data and learns a separate exploration policy, whereas our approach seeks to directly fine-tune the pre-trained policy itself. 
We compare the performance of PORL with JSRL under BC pre-training setting (see Fig.~\ref{fig:bc_setting}) and offline-to-online RL setting (see Fig.~\ref{fig:jsrl}).
JSRL demonstrates unstable performance in complex tasks (e.g., \texttt{Antmaze-Large}), as its curriculum learning mechanism necessitates manual scheduling for challenging environments.

\section{When should PORL not be used?}
\label{relocate}
\begin{figure}[htbp]
    \centering
    \includegraphics[width=0.6\linewidth]{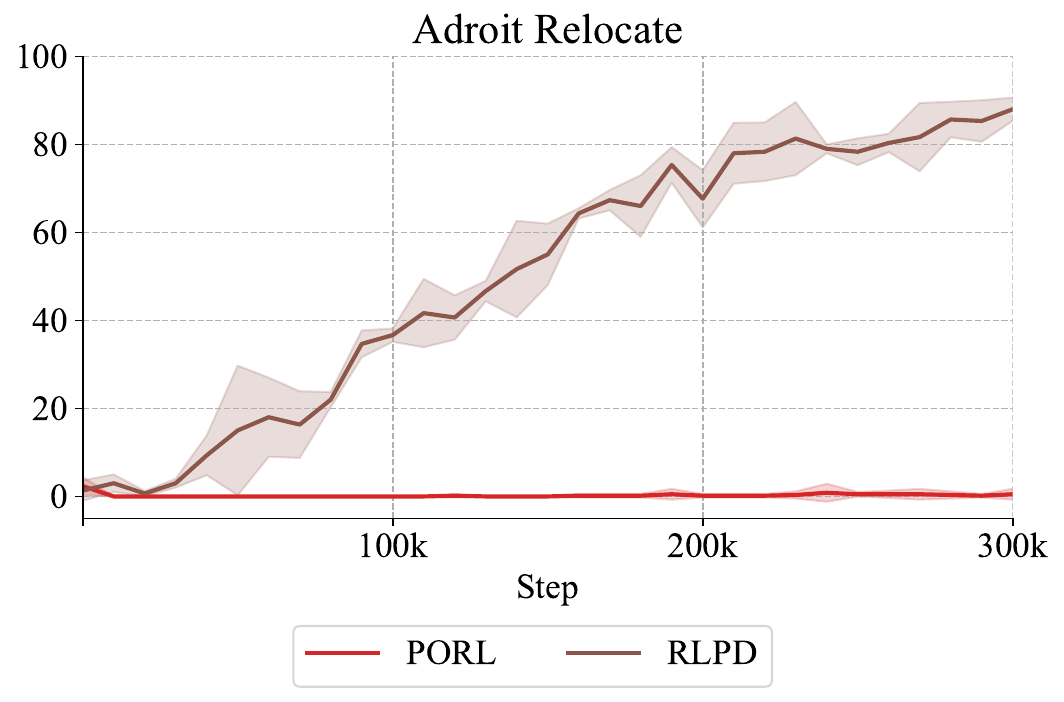}
    \caption{PORL does not work when the performance of pretrained policies is poor.}
    \label{fig:bad_case}
\end{figure}

PORL serves as an online RL fine-tuning method for pretrained policies. When the performance of pretrained policies is very weak, there is no need to use PORL instead of online methods like RLPD (e.g., \texttt{Adroit-Relocate} and \texttt{Kitchen-complete} in Fig.~\ref{fig:bad_case}).

\section{The effect of offline data retention}
\begin{figure}[htbp]
    \centering
    \includegraphics[width=1\linewidth]{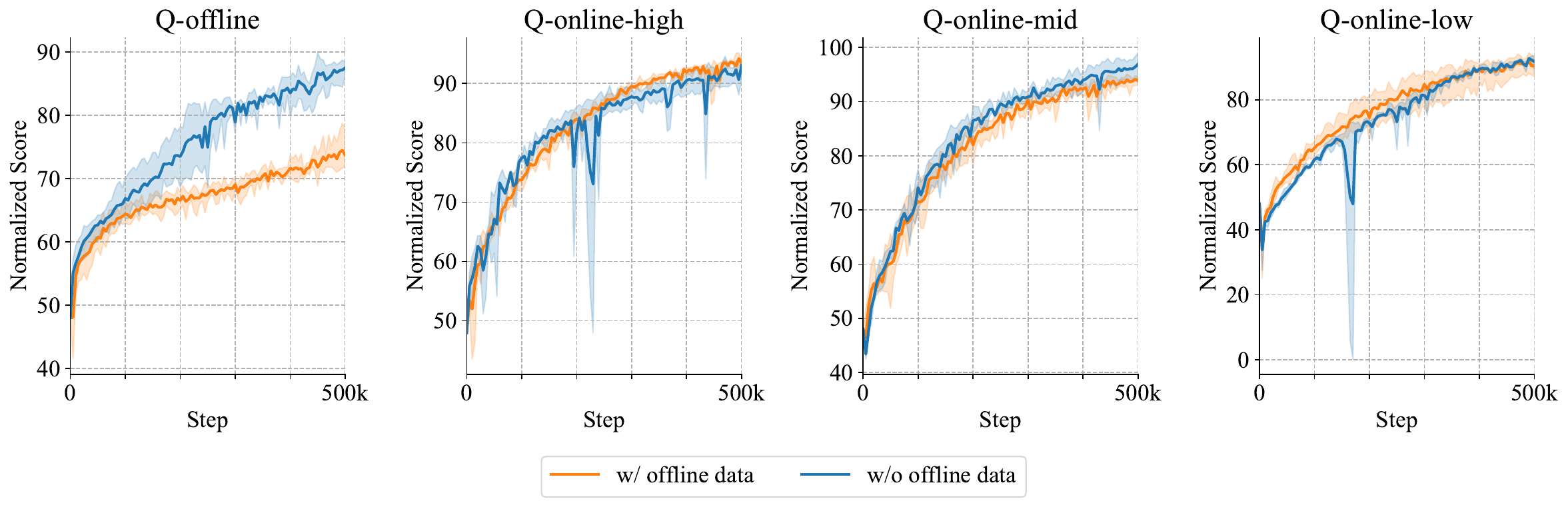}
    \caption{Comparison of online RL fine-tuning pre-trained policy with or without retaining offline data.}
    \label{fig:Q_tuning_wo_off_data}
\end{figure}

The experiments in Sec.~\ref{sec3} adhere to the common setting of symmetric sampling of offline and online data buffers in previous research. 
To conduct a more comprehensive exploration, we investigate online fine-tuning where offline datasets are not retained, as illustrated in Figure \ref{fig:Q_tuning_wo_off_data}.
Compared to retaining offline data, the approach of not retaining offline data generally results in improved average normalized scores, accompanied by increasing variance.
This suggests that policy learning is accelerated but less stable. 
These results align with the perspectives of previous research \cite{zhou2024efficient}, indicating that while offline data retention tends to ensure the stability of online fine-tuning, it may conversely hinder efficiency and asymptotic performance.

\section{Implement details}\label{a2}

\subsection{Baselines}
\label{baseline}
\begin{itemize}
\item \textbf{CQL} \cite{kumar2020conservative} is an offline RL method that regularizes value estimates to prevent overestimation. 
\item \textbf{IQL} \cite{kostrikov2021offline} is an offline reinforcement learning algorithm that optimizes policies without explicitly relying on Q-value estimates, using an implicit approach to avoid overestimation and enhance stability.
\item \textbf{Cal-QL} \cite{nakamoto2024cal} is a state-of-the-art offline-to-online RL algorithm that calibrates the Q-function in CQL for efficient fine-tuning. 
\item \textbf{WSRL} \cite{zhou2024efficient} is a state-of-the-art offline-to-online RL algorithm that does not retain offline data and introduces a warm-up phase to bridge offline pre-training and online tuning. 
\item \textbf{RLPD} \cite{ball2023efficient} is a state-of-the-art online RL algorithm with offline data that leverages symmetric sampling of the offline dataset and online buffer. 
\item \textbf{JSRL} \cite{uchendu2023jump} is a curriculum learning-based online RL algorithm that utilizes a pre-trained policy as a prior. In our experiments, the pre-trained policy serves as the policy prior. 
\end{itemize}

All baseline implementations leverage publicly available codebases with default hyperparameters to ensure reproducibility:  
\begin{itemize}
\item CQL, IQL: Implemented via the CORL framework \citep{tarasov2022corl} (\url{https://github.com/tinkoff-ai/CORL}), which provides standardized offline RL algorithms. 
\item WSRL: \url{https://github.com/zhouzypaul/wsrl}. PORL and JSRL implementations build upon the WSRL codebase. 
\item Cal-QL: \url{https://github.com/nakamotoo/Cal-QL}. 
\item RLPD: \url{https://github.com/ikostrikov/rlpd}. 
\end{itemize}

WSRL, RLPD, and JSRL take the same sample efficiency technicals as ours, including LayerNorm, high UTD, and Q-ensemble as the original implementation to ensure a fair comparison.
The results in Fig.~\ref{fig:policy_only} and Fig.~\ref{fig:off2on} are averaged over six seeds, and the shaded area represents standard deviations.
The results in ablations are averaged over three seeds, and the shaded area represents the min-max interval.
\subsection{Benchmarks}
\label{benchmark}

\subsubsection{AntMaze}
\begin{figure}[htbp]
    \centering
    \includegraphics[width=0.8\linewidth]{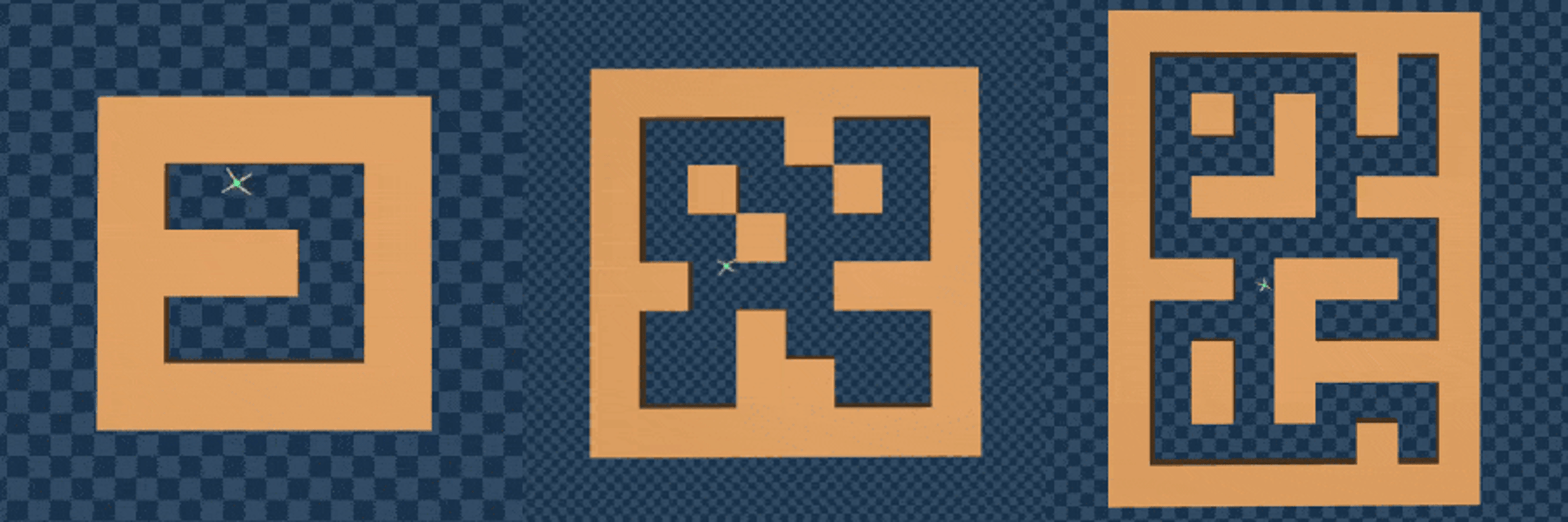}
    \caption{Antmaze tasks visualizations (Umaze, Medium, and Large).}
    \label{fig:antmaze_visual}
\end{figure}
\begin{itemize}
    \item \textbf{Overall Description}:  
    A navigation benchmark controlling an 8-DoF quadruped "Ant" robot to move from a starting point to a goal in procedurally generated mazes. The domain mimics real-world robotic navigation by replacing simple agents (e.g., Maze2D’s 2D ball) with a morphologically complex robot (see Fig.~\ref{fig:antmaze_visual}).
    
    \item \textbf{Tasks Description}: Three difficulty levels: \textit{Umaze} (simple), \textit{Medium}, and \textit{Large} (complex branching paths).
    \textit{Fixed Goal}: antmaze-umaze-v2 (fixed start-to-goal pairs).
    \textit{Diverse}: Random start locations and random goals (e.g., medium-diverse, large-diverse).
    \textit{Play}: Trajectories guided to specific hand-picked intermediate locations (not goals), from curated start points (e.g., medium-play, large-play).
    
    \item \textbf{Feature}:  
    \textit{Sparse Rewards}: Binary rewards (0/1) only reaching the goal, with no intermediate rewards.
    \textit{Morphological Complexity}: 8-DoF Ant dynamics increase real-world relevance compared to simpler agents;
    \textit{Non-Markovian Policies}: Policies rely on historical waypoint tracking, breaking Markov assumptions;
\end{itemize}

\subsubsection{Adroit}
\begin{figure}[htbp]
    \centering
    \includegraphics[width=0.8\linewidth]{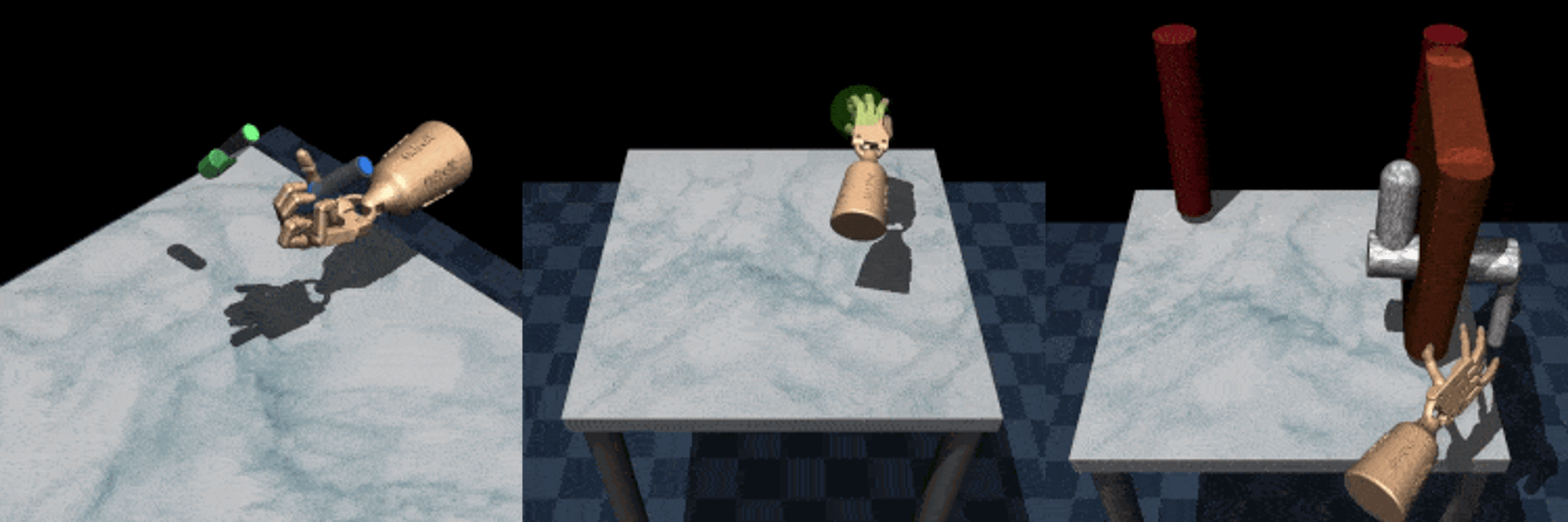}
    \caption{Adroit tasks visualizations (pen, relocate, and door).}
    \label{fig:adroit_visual}
\end{figure}
\begin{itemize}
    \item \textbf{Overall Description}:  
    A dexterous manipulation benchmark controlling a 24-DoF Shadow Hand robot to perform complex object interaction tasks. It focuses on learning from narrow expert datasets with high-dimensional action spaces and sparse rewards, simulating real-world robotic manipulation challenges (see Fig.~\ref{fig:adroit_visual}).

    \item \textbf{Tasks Description}: Three manipulation tasks: 
    \textit{Pen}: Rotate a pen to target orientation;
    \textit{Relocate}: Move an object to target location. 
    \textit{Door}: Open a door with a latch; 

    \item \textbf{Feature}:  
    \textit{High-Dimensional Control}: 24-DoF hand dynamics requiring precise coordination;
    \textit{Sparse Rewards}: Binary rewards (0/1) only upon task completion;
    \textit{Narrow Data Distribution}: Limited expert demonstrations and action space complexity;
\end{itemize}

\subsubsection{Kitchen}
\begin{figure}[htbp]
    \centering
    \includegraphics[width=1\linewidth]{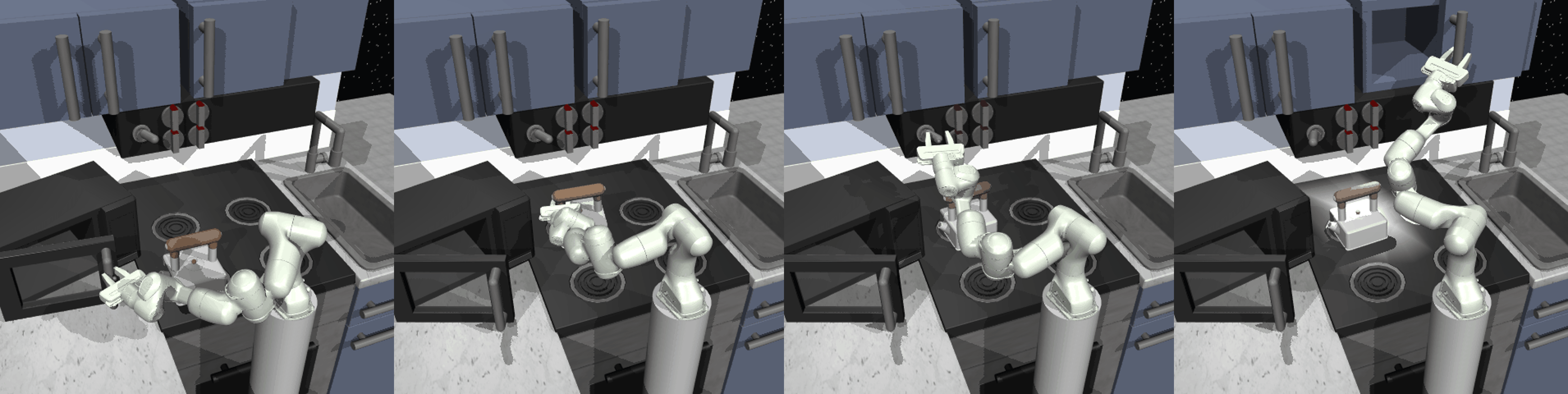}
    \caption{Kitchen tasks visualizations.}
    \label{fig:kitchen_visual}
\end{figure}
\begin{itemize}
    \item \textbf{Overall Description}:  
    A multi-stage manipulation benchmark controlling a 9-DoF Franka robot to arrange a kitchen environment into a desired configuration. It requires solving 4 subtasks in sequence, testing combinatorial task planning and long-horizon policy learning under sparse rewards (see Fig.~\ref{fig:kitchen_visual}).

    \item \textbf{Tasks Description}: 
    \textit{Subtasks}: Sequential completion of 4 kitchen arrangement actions (e.g., opening microwaves, moving objects).  
    \textit{Datasets}: Three data optimality levels:  
    - \textit{Complete}: Full task demonstrations;  
    - \textit{Partial}: Mix of complete/incomplete subtask trajectories;  
    - \textit{Mixed}: Only incomplete trajectories.  

    \item \textbf{Feature}:  
    \textit{Compositional Rewards}: Incremental rewards (0 to +4) based on completed subtask count;  
    \textit{Multi-Stage Planning}: Requires learning subtask execution order;  
\end{itemize}

\subsubsection{RLBench}
\begin{figure}[htbp]
    \centering
    \includegraphics[width=0.8\linewidth]{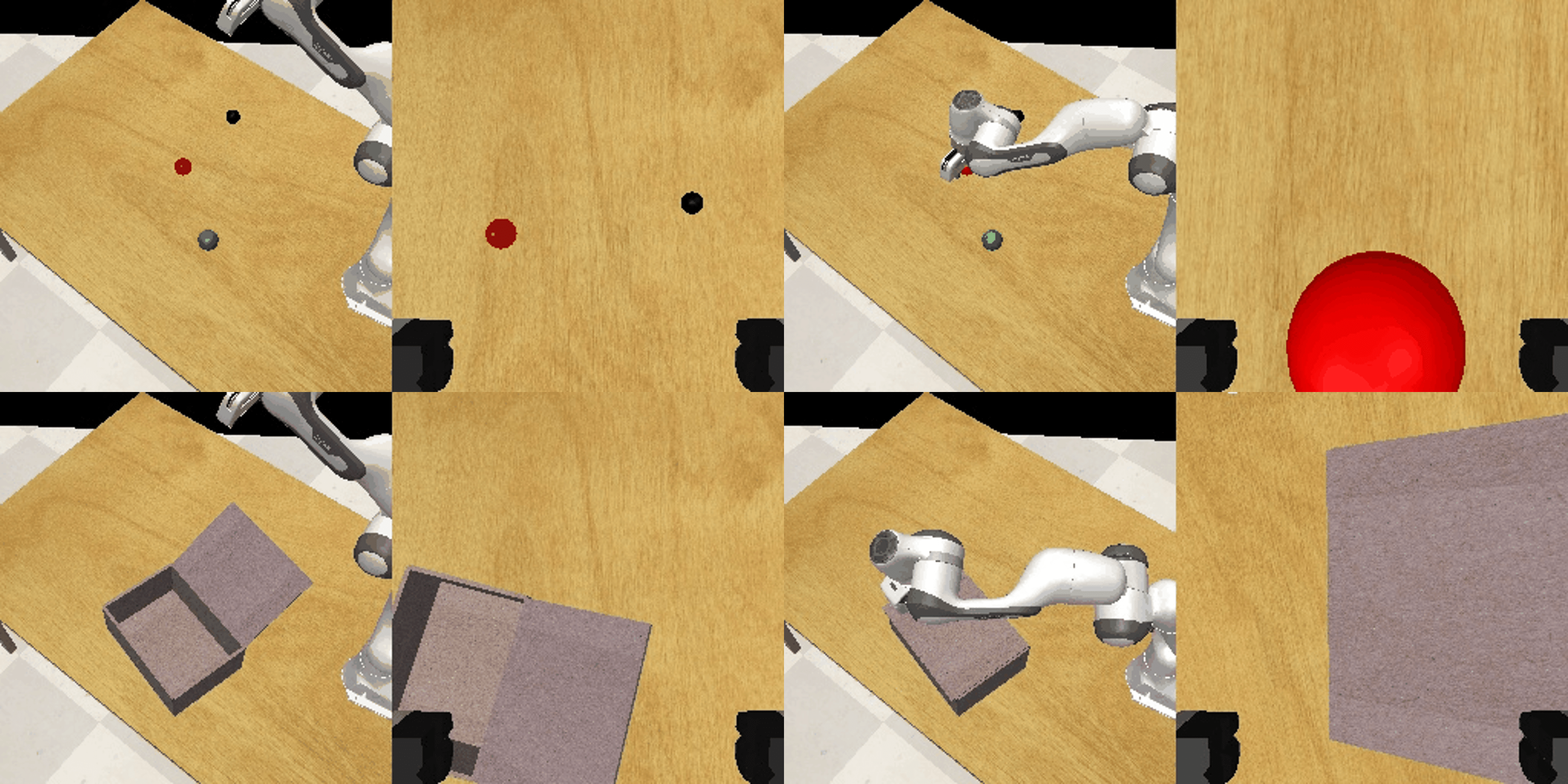}
    \caption{RLBench tasks visualizations.}
    \label{fig:rlbench_visual}
\end{figure}
Below are some additional details regarding the RLBench \citep{james2019rlbench} experimental setup:

\begin{itemize}
    \item \textbf{ReachTarget} (grasping the red ball, see top in Fig.~\ref{fig:rlbench_visual}), the input state dimension is 10 (7 joint positions + 3 task low-dimensional state), and the action dimension is 8 (7 joint velocities + 1 gripper open/close). The total terminal steps are set to 70.
    \item \textbf{Closebox} (grasping the lid of a box with the robot's gripper and closing the lid, see bottom in Fig.~\ref{fig:rlbench_visual}), the input state dimension is 64 (7 joint positions + 57 task low-dimensional state), and the action dimension is 8 (7 joint velocities + 1 gripper open/close). The total terminal steps are set to 270.
\end{itemize}

\subsection{Hyperparameter}
The hyperparameter used in our experiments is detailed in Table~\ref{tab:hyperparameter}.

\begin{table}[htbp]
    \centering
    \caption{Hyperparameter of All Tasks.}
    \resizebox{1.0\linewidth}{!}{
    \begin{tabular}{l|cccc}
    \toprule 
    \textbf{Hyperparameter}            & \textbf{Antmaze} & \textbf{Adroit} & \textbf{Kitchen} & \textbf{RLBench} \\ 
    \midrule
    \textbf{PORL}                  &                  &                 &                  \\ 
    \midrule
    Epsilon                   & 0.1              & 0.1             & 0.1             & 0.1              \\ 
    Pre-sample steps          & 20k              & 5k              & 20k              & 20k              \\ \midrule
    \textbf{Optimization}          &                  &                 &                 &                  \\ \midrule
    Actor Learning Rate       & $1e^{-4}$        & $1e^{-4}$       & $1e^{-4}$       & $1e^{-4}$        \\ 
    Critic Learning Rate      & $3e^{-4}$        & $3e^{-4}$       & $3e^{-4}$       & $3e^{-4}$        \\ 
    Temperature Learning Rate & $1e^{-4}$        & $1e^{-4}$       & $1e^{-4}$       & $1e^{-4}$        \\ 
    Batch Size                & 512              & 512             & 1024             & 1024              \\ 
    Optimizer                 & Adam             & Adam            & Adam            & Adam             \\ 
    UTD                       & 16               & 16              & 16              & 16               \\ 
    Offline Steps             & 1000,000         & 40,000          & 250,000          & -                \\ \midrule
    \textbf{Architecture}          &                  &                 &                  \\ \midrule
    Actor Network             & [256, 256]       & [512, 512] & [512, 512, 512]      & [256, 256]       \\ 
    Critic Network            & [256, 256, 256, 256] & [512, 512, 512] & [512, 512, 512] & [256, 256, 256, 256] \\ 
    Q-ensemble                & 10               & 10              & 10              & 10               \\ 
    Activations               & Relu             & Relu            & Relu            & Relu             \\ \bottomrule
    \end{tabular}
    }
\label{tab:hyperparameter}
\end{table}

\end{document}